\documentclass[runningheads]{llncs}
\usepackage{graphicx}
\usepackage[utf8]{inputenc} 
\usepackage{hyperref}       
\usepackage{url}            
\usepackage{booktabs}       
\usepackage{amsfonts}       
\usepackage{nicefrac}       
\usepackage{microtype}      
\usepackage{lipsum}
\usepackage{graphicx}
\usepackage{epstopdf}
\usepackage[ruled,vlined]{algorithm2e}
\usepackage{subcaption}
\usepackage[T1]{fontenc}
\usepackage{lmodern}

\begin{document}
\title{Curious Exploration and Return-based Memory \\Restoration for Deep Reinforcement Learning
}
%
%
\author{Saeed Tafazzol \inst{1} \and
Erfan Fathi \inst{1} \and
Mahdi Rezaei \inst{2} \href{https://orcid.org/0000-0003-3892-421X}{\includegraphics[width=3mm]{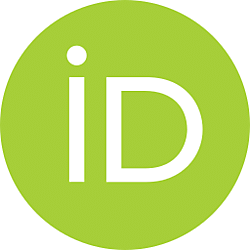}}
\and
Ehsan Asali \inst{3}}
\authorrunning{S. Tafazzol et al.}
%
\institute{Advanced Mechatronics Researcher Lab, IR. \, \email{\{s.tafazzol,e.fathi\}@qiau.ac.ir} \and
Institute for Transport Studies, University of Leeds, UK. \, \email{m.rezaei@leeds.ac.uk} \and
The University of Georgia, USA. 
\,\email{ehsanasali@uga.edu}\\
}
\maketitle              
\begin{abstract}
Reward engineering and designing an incentive reward function are non-trivial tasks to train agents in complex environments. 
Furthermore, an inaccurate reward function may lead to a biased behaviour which is far from an efficient and optimised behaviour. In this paper, we focus on training a single agent to score goals with binary success/failure reward function in Half Field Offense domain. As the major advantage of this research, the agent has no presumption about the environment which means it only follows the original formulation of reinforcement learning agents. The main challenge of using such a reward function is the high sparsity of positive reward signals. To address this problem, we use a simple prediction-based exploration strategy (called Curious Exploration) along with a Return-based Memory Restoration (RMR) technique which tends to remember more valuable memories. The proposed method can be utilized to train agents in environments with fairly complex state and action spaces. Our experimental results show that many recent solutions including our baseline method fail to learn and perform in complex soccer domain. However, the proposed method can converge easily to the nearly optimal behaviour. 
The video presenting the performance of our trained agent is available at \href{http://bit.ly/HFO_Binary_Reward}{http://bit.ly/HFO\_Binary\_Reward}.

\end{abstract}
\section{Introduction}
Machine learning is one of the main sub-categories of AI with applications in various domains such as healthcare \cite{rezaei2020deepsocial}, 
autonomous vehicles \cite{das2010} or robotics systems \cite{asali2018namira}. The general objective is to teach single or multiple agents to successfully perform a task with a minimum guidance from a human. Reinforcement Learning is a leading paradigm for teaching hard-to-code tasks to robotic agents. A reinforcement learning agent mainly relies on the experience it gains through some trial and error runs. Such an agent focuses on finding a balance between exploration and exploitation and it tries to learn actions with long-term benefits. In other words, the agent repeatedly interacts with an unknown environment with the goal of maximising its cumulative reward \cite{sutton1998introduction}.
%
%
To enable the agent to perceive rewards from the environment, a human expert has to design a reward function specific to the domain which is subjective to the expert's knowledge about the task. Moreover, the hand-crafted reward function may compromise the goal we truly care about. To address the difficulty with the design of a decent reward function for a reinforcement learning agent, this paper focuses on a problem mentioned in  \cite{hausknecht2015deep} namely, reward engineering. M. Hausknecht and P. Stone propose a method capable of training a single agent to learn how to score on an empty goal from scratch, based on a hand-crafted reward function. Without this reward function, the agent is hopeless to learn anything. In order to overcome such a limitation, we suggest a prediction-based exploration strategy (namely, curious explorer) which is much more successful than a mere random action selection algorithm. Using curious exploration, the agent attempts to perform actions which it has less clue about its outcome (this behaviour is called ``curiosity'' in animals \cite{berlyne1966curiosity}). This way, the agent is always curious about novel situations while learning from previous experiences. In other words, the exploration will be highly affected by the curiosity of the agent, leading to a better convergence rate to the nearly-optimal behaviour. According to our experiments, the agent will eventually score goals. 
Our proposed exploration strategy turns the problem of absent positive reward signals into a sparse one. To increase the density of positive reward signals in replay memory, we utilise a different replay memory than the standard one, in which seeks to remember more valuable memories rather than bleak ones. 

One of the main contributions of this work is taking advantage of a binary reward function which directly focuses on the task's main objective. 
Furthermore, note that our work only follows the standard formulation of RL which only relies on the next state as well as reward signal, both coming from the environment.

We use RoboCup 2D Half Field Offense (HFO) domain \cite{hausknecht2016half} to perform our experiments. Using the proposed method in \cite{hausknecht2015deep} as our baseline, we focus on the task of scoring on an empty goal with a binary success/failure reward function. However, instead of a hand-crafted reward signal, we suggest using a binary reward signal. To deal with the challenges in exploiting sparse binary reward signals, we suggest using Curious Exploration as well as Return-based Memory Restoration. The experimental results show that after experiencing sufficiently enough game episodes, the agent can reliably score goals almost every time. 

The rest of this paper is organised as follows: the related work is discussed in Section 2; Section 3 explains the background on deep reinforcement learning in parametrised action space and the HFO domain. Section 4 presents our proposed method and finally, Section 6 shows and analyses the experimental results followed by future work and conclusion.

\section{Related Work}

Experience replay memory \cite{lin1993reinforcement} plays a vital role in the performance of Deep RL algorithms. Many advancements have been made to improve the efficiency of this component. For instance, Hindsight Experience Replay (HER) \cite{andrychowicz2017hindsight} saves all the state transitions of an episode in the replay buffer not only with the original goal of the episode, but also with a subset of other goals. 
The work uses a binary reward signal and can converge fast to the solution. HER is explicitly designed to handle binary reward which has been rarely used in other works. Nonetheless, it is hard if not impossible to implement HER in adversarial situations. 

Another work called Prioritized Experience Replay (PER) \cite{schaul2015prioritized} that modifies the sampling process of the experience replay memory. PER relies on Temporal Difference (TD) error to prioritise over the memories in the replay buffer. 
Handling a replay memory involves two decisions; which experiences to store, and which experiences to replay. PER addresses only the latter while our proposed method focuses on the former.

Intrinsic Curiosity Module (ICM) \cite{pathak2017curiosity} is an exploration method that learns the state space encoding with a self-supervised inverse dynamics model. ICM wisely chooses the features of the state space; i.e., it only considers the features that can influence the agent's behaviour and ignores others. Having the ability to ignore the state factors that the agent has no incentive to learn them enables ICM to solve the noisy-TV problem. Since our environment does not suffer from noisy-TV problem and also the state space prediction in 2D soccer simulation environment is not very complex, in our work, we adhere to the original version of Curious Exploration.

Zare et al. \cite{zarecyrus} have implemented a goalie cooperating defender’s decision making in HFO domain using Deep Deterministic Policy Gradient (DDPG). Their work exploits binary success/failure reward function and predefined hand-coded actions (so called ``high-level actions''). Having used the high-level behaviours, the ratio of positive rewards is high enough to learn the task. In our case, we use the original HFO actions (so called ``low-level actions''). This way, the agent is not prone to subjective implementation of hand-coded actions and learns the task from scratch, objectively.

\section{Background}


In this section we provide the fundamental background about Deep Deterministic Policy Gradient (DDPG) \cite{lillicrap2015continuous} (a robust deep reinforcement learning algorithm) and Half Field Offense Domain \cite{hausknecht2016half} as the prerequisite for the section Methodology to control our agent(s) in continuous action space of the soccer robot field.

\subsection{Deep Deterministic Policy Gradient}
\begin{figure}[t!]
    \centering
    \vspace{-3mm}
    \includegraphics[width=7cm]{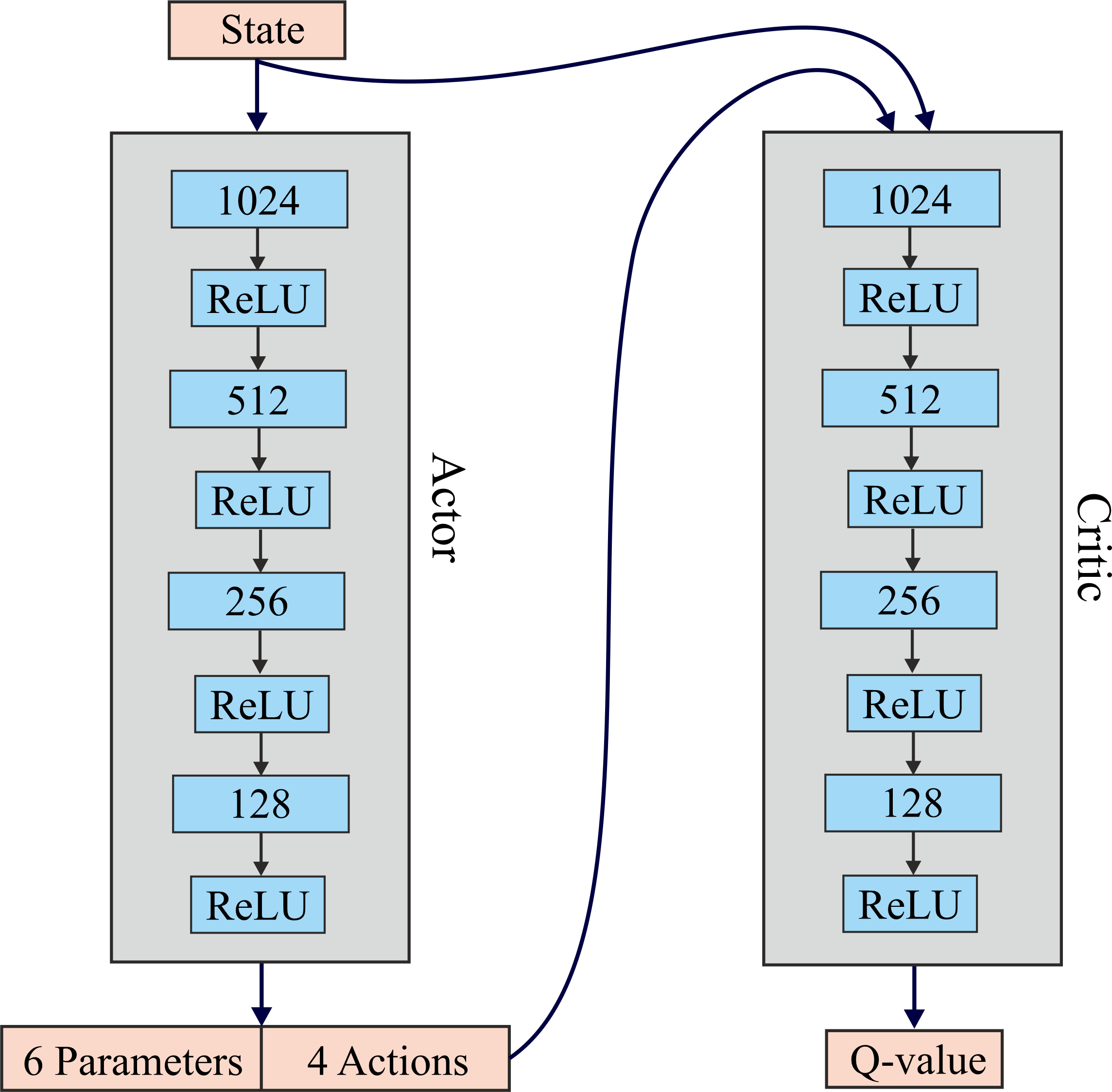}
    \vspace{-2mm}
    \caption{Actor/Critic network architecture used in parametrised action space. Note that along with continuous parameters, four discrete scalars are used to indicate which action is active \cite{hausknecht2015deep}}
    \label{fig:ddpg}
\end{figure}

This method uses Actor/Critic architecture which decouples the value learning and action selection (Fig. \ref{fig:ddpg}). Actor network \(\mu\) parametrised with \(\theta_\mu\) gets the \textit{state} as input in order to output the desired action. Critic network \(Q\) parametrised with \(\theta_Q\) gets the \textsl{state} and \textit{action} as input, then estimates the Q-value for the input action. To update Q-value the critic uses Temporal Difference (TD) \cite{tesauro1995temporal} equation which defines a critic loss function for neural network setting as follows: 
\begin{equation}
	L_Q(s,a|\theta^{Q}) = (Q(s,a|\theta^{Q}) - (r + \gamma{Q}(s',\mu(s'|\theta^{\mu})'|\theta^{Q})))^{2}
\end{equation}
On the other hand, the loss function for the actor model is simply \(-Q\) which yields to the following gradient with respect to the actor's parameters: 
\begin{equation}
-\nabla_{\theta^\mu}Q(s,a|\theta^Q) = -\nabla_a{Q}(s,a|\theta^Q)\nabla_{\theta^\mu}\mu(s|\theta^\mu)
\end{equation}
A Parametrised Action Space Markov Decision Process (PAMDP) \cite{masson2016reinforcement} is defined by a set of discrete actions, \(A_d=\{a_1,a_2,...,a_k\}\). Each discrete action \(a \in A_d \) features \(m_a\) continuous parameters, \( \{p^a_1,p^a_2,...,p^a_k\} \in \mathbb{R}^{m_a}\). In this process, an action must be chosen first; then, paired with associated parameters. Soccer 2D simulation \cite{akiyama2010agent2d,akiyama2013helios} environment is modelled as a PAMDP with four actions including dash, turn, kick, and tackle while each action has its own parameters. 



To take advantage of DDPG, Hausknecht and Stone \cite{hausknecht2015deep} use a scalar for each discrete action to determine which action must be chosen. Along with the parameters for each action, this constitutes a continuous action space (Fig.\ref{fig:ddpg}). For example, four discrete scalars for the actions (dash, turn, kick, and tackle) are paired with their parameters (a total of 6 parameters) in HFO domain. 
%
As mentioned before, DDPG's critic network utilises TD formula to update the Q-value function. However, there is also an alternative option for updating the Q-value function, called Monte Carlo (MC) \cite{sutton1998introduction}, which is able to learn online from experience. Nonetheless, opposed to TD-methods, Monte Carlo approaches do not bootstrap value estimates and instead learn directly from returns. In other words, On-policy MC employs on-policy updates without any bootstrapping, while Q-Learning
uses off-policy updates with bootstrapping. Both TD and MC try to roughly calculate the action-value function \(Q(s, a)\) directly from experience tuples of the form \((s_t, a_t, r_t, s_{t+1}, a_{t+1})\). Furthermore, having visited all state-value pairs an infinite number of times, both methods can provably converge to optimally. Nevertheless, the fundamental contrast between the two methods can be realised by investigating their update targets. The update target formula for Monte Carlo is defined as follows: 
\begin{equation}
	\hat{R} = \sum_{i=t}^{T} \gamma^{i-t}r_i
\end{equation}
\noindent and the update target function for Temporal Difference is defined as:
\begin{equation}
q_{target} = r + \gamma{Q}(s',\mu(s'|\theta^{\mu})'|\theta^{Q})
\end{equation}
To stabilise the learning procedure, \cite{hausknecht2016policy} proposes a mixed update paradigm for critic network in the following form:
\begin{equation}
y = \beta{\hat{R}} + (1-\beta){q_{target}}	
\end{equation}
\noindent where \(\beta\) is a scalar \(\in[0,1]\) indicating the mixture of returned reward in the episode (\(\hat{R}\)) and the target value for q-learning (\(q_{target}\)). In our experiments, we use \(\beta = 0.2\). According to the definition of \(y\), we modify the loss function for the mixture of the two algorithms as follows:
\begin{equation}
L_Q(s,a|\theta^{Q}) = (Q(s,a|\theta^{Q}) - y)^{2}\	
\end{equation}



\subsection{Half Field Offense Domain}
Half Field Offense (HFO) \cite{hausknecht2016half} domain is a simplified soccer 2D simulation task where agents attempt to defend or score goals. 
HFO features a continuous state space and a parametrised action space where we will explore more in the following sections.
%
%
In the context of State Space, the agent uses egocentric continuously-valued features. These features include the angle and the distance to important objects in the field such as ball, players, and landmarks. As mentioned before, HFO features a parametrised action space. The mutually-exclusive discrete actions to choose from are as follows:
\begin{enumerate}
	\item \(DASH(power,direction)\) moves in the indicated direction \(\in [-180,180]\) with power \(\in [0,100]\). It must be noted that agent moves faster in the direction that aligns with its body angle compared to other directions.
	\item \(TURN(direction)\) turns according to the indicated direction \(\in [-180,180]\) 
	\item \(KICK(power,direction)\) kicks the ball if the agent is kickable according to power \(\in [0,100]\)  and direction \(\in [-180,180]\).
	\item \(Tackle(power)\) tackles the ball or other players with power \(\in [0,100]\). Note that tackle is usually effective in defensive behaviour and we do not use it in our experiments since we train agents for offensive behaviour.
\end{enumerate}
The reward signal used in \cite{hausknecht2015deep} is hand-crafted in order to guide the agent through the process of scoring a goal. The reward signal in their work is defied as follows:
\begin{equation}
	r_t = d_{t-1}(a,b) - d_t(a,b) + \mathbb{I}_t^{kick} + 3(d_{t-1}(b,g) - d_t(b,g)) + 5\mathbb{I}_t^{goal}
\end{equation}
where \(d_t(a,b) , d_t(b,g)\) are respectively the distance of the agent from the ball and the distance of the ball from the goal centre at time \(t\). 
\(\mathbb{I}_t^{kick}\) is a binary variable indicating that the agent has touched the ball for the first time at time \(t\).\(\mathbb{I}_t^{goal}\) is also a binary variable indicating that the agent scored a goal at time \(t\).\\
However, for our research we only use the last term or more precisely: 
\begin{equation}
r_t = 5\mathbb{I}_t^{goal}	
\end{equation}
We show that our method, unlike the one in \cite{hausknecht2015deep}, is able to learn the task of scoring goals with this new reward signal formulation.

\section{Proposed Method}
Reinforcement learning is hopeless to learn anything without encountering a desirable reward signal in exploration phase \cite{sutton1998introduction}. In our case, due to the fact that a random agent never scores a goal, by using a success/failure reward signal, our baseline article \cite{hausknecht2015deep} fails to learn the task. To overcome this issue, we use a prediction-based exploration strategy called ``Curious Exploration''. However, this exploration strategy degrades the problem of completely absent positive reward signal into a sparse one. To deal with the sparsity, we modify the replay memory in a way that it tends to remember more promising memories and we call it ``Return-based Memory Restoration''. Complete source code for our agent is available at \href{https://github.com/SaeedTafazzol/curious_explorer}{https://github.com/SaeedTafazzol/curious\_explorer}. In the following subsections, we will describe the two mentioned approaches in more detail.

\subsection{Exploration and Exploitation Agents}

Our framework takes advantage of two internal agents that take responsibility to perform either the exploration or the exploitation task. Indeed, one agent only focuses on the exploration by looking for novel states while the other agent concentrates on exploitation attempting to score goals. However, since only one of these two agents can be active at a time, the soccer agent has to follow a procedure to switch between the two. 
A widely used policy to obtain a satisfactory trade-off between exploration and exploitation is \(\epsilon-greedy\) \cite{sutton1998introduction} where \(\epsilon\) indicates the probability of taking a random action. In our framework, \(\epsilon\) is the probability of activating the exploration agent. Hausknecht and Stone \cite{hausknecht2015deep} anneal \(\epsilon\) from 1.0 to 0.1 over the first 10,000 updates in the experiments. Nevertheless, this annealing function only works well with a hand-crafted incentive reward function. Regarding the fact that our reward signal is a binary function which is quiet different from our baseline's reward function, we use a different annealing function which anneals the amount of exploration by a constant factor every time a positive reward signal is gained. This choice of annealing function relies on the fact that a reinforcement learning agent cannot learn anything with a dead reward signal. In other words, it is pointless to expect the agent to perform exploitation when it has not seen any reward signal other than zero.

\subsection{Curious Exploration (CE)}
Exploration in reinforcement learning assists the agent to observe new patterns which can lead to a better learned policy \cite{sutton1998introduction}. Random action selection is a commonly used exploration approach; however, it is highly probable to visit previously seen states while exploration. Therefore, exploring big environments is problematic using random selection. As an example, a random moving agent in the space has the expectation of staying at the same place.
Therefore, to visit more novel states, we should design a proper exploration strategy. 

The first challenge to solve is to come up with a method to identify and differentiate the novel states from previously seen states in continuous state-space. One potential solution is to have a model responsible for learning the forward dynamics of the environment. This way, if the model can predict the future state, we can assume that the agent has previously been in a similar situation and vice versa. Note that the prediction accuracy is a good metric to examine the performance of the learned forward dynamics model. To formulate the learning procedure, we exploit the formulation of Markov Decision Process for reinforcement learning. To elaborate more on this, lets assume the agent is at \(s_t\) attempting to take the action \(a_t\). Performing the given action, the agent's predictor should be able to identify \(s'_{t+1}\) and \(r'_{t+1}\) as the foreseen values for \(s_{t+1}\) (the next state) and \(r_{t+1}\) (the upcoming reward), respectively. The predictor \(P\) then has the following loss function:
\begin{equation}
L_P = {||(P(s_t,a_t) - (s_{t+1},r_{t+1}))||}^2
\end{equation}
where \((s_{t+1},r_{t+1})\) is the concatenation of the next state and the upcoming reward vectors. Such a loss function formulation, which is only dependent to the transitions, enables the predictor to learn the model in a supervised manner.
\begin{figure}[t!]
    \centering
    \vspace{-2mm}
    \includegraphics[width=10cm]{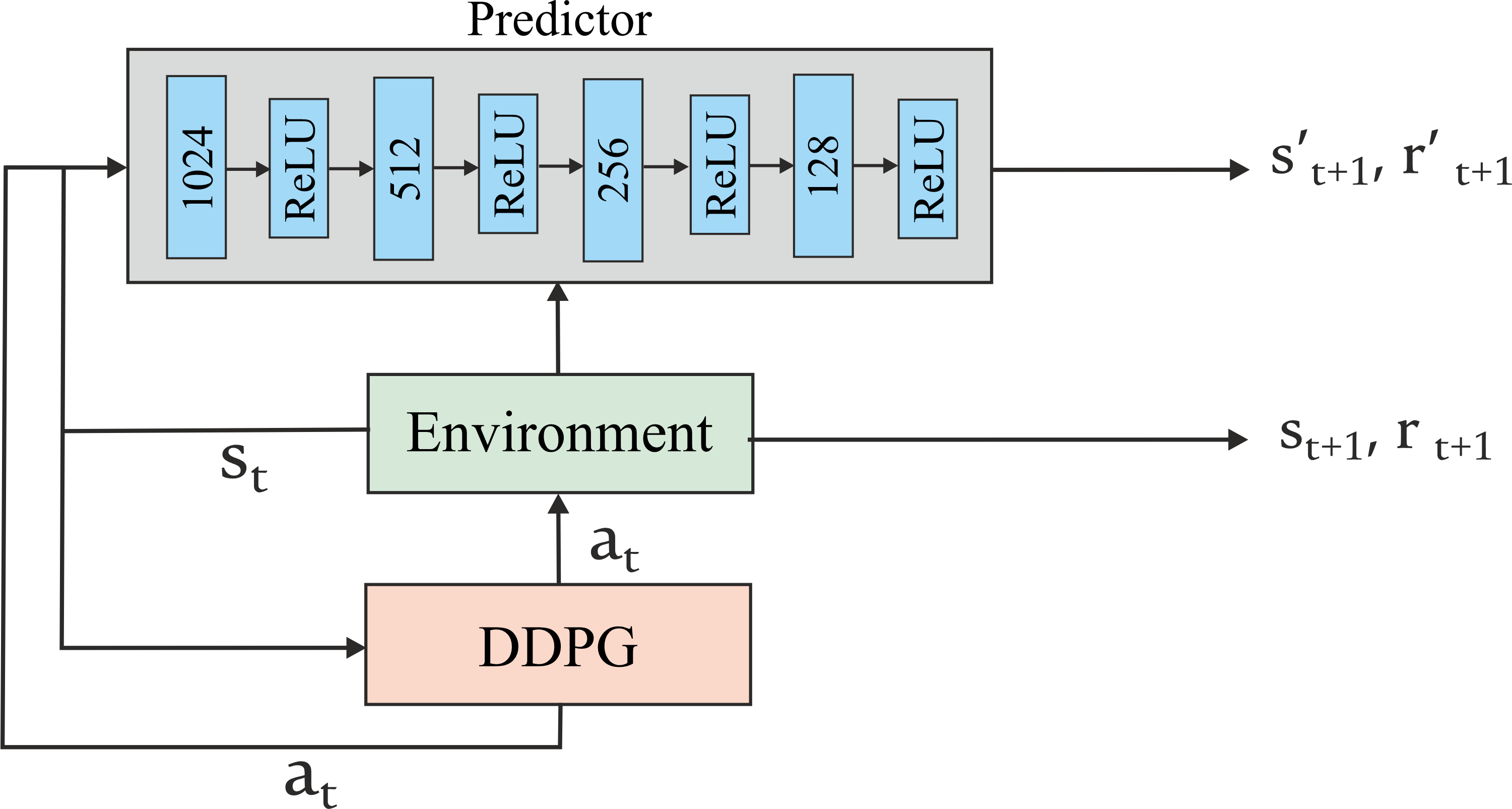}
    \vspace{-2mm}
    \caption{The overall structure of Curious Exploration (CE) algorithm}
		\vspace{-2mm}
    \label{fig:ce}
\end{figure}
Having an evaluation metric to measure the status novelty \((L_P)\), we can now add the exploration component to our framework. To do so, we can use the predictor's loss function as a reward signal (which is called ``intrinsic reward'') for an explorer agent (Fig.\ref{fig:ce}).
Opposed to the prediction model, which is a regression problem, reaching a novel state may require the agent to perform a series of actions (not necessarily one action). This means that the exploration process is basically a reinforcement learning problem that tends to reach the desired outcome in the long run. Having that in mind, we can calculate intrinsic reward in the following form:
\begin{equation}
r_t^i = L_P	
\end{equation}
Interestingly enough, animals use the same notion to explore their world more efficiently. More precisely, they like to examine things which they are unsure of the outcome; this intention in animals is called ``Curiosity''  \cite{schmidhuber1991possibility}. The reason this method closely follows the curiosity behaviour is that our agent tries to perform actions with unknown outcomes. We show that this method guides the agent to score goals during the training process which leads to generate numerous promising memories. The experimental results show that such amount of promising memories is sufficient for the agent to learn how to score goals.
\subsection{Return-based Memory Restoration (RMR)}
The utilisation of curious exploration may lead to some memories with positive reward signals; however, the memories may be too sparse for the agent to properly learn a difficult task in complex environments. For example, the soccer agent may only score a goal every few episodes.

\begin{figure}[t!]
    \centering
    \vspace{-1mm}
    \includegraphics[width=12cm]{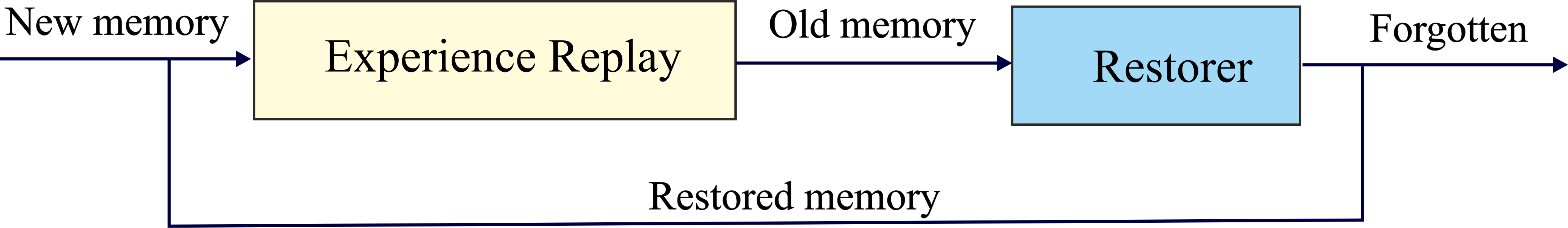}
    \vspace{-1mm}
    \caption{The architecture of Return-based Memory Restoration (RMR)}
    \vspace{-2mm}
    \label{fig:rmr}
\end{figure}

In order to deal with the sparsity of the memories with positive rewards, we modify the structure of the replay memory to motivate it to remember more valuable memories. To do so, one potential solution is to use the reward saved in each transition as an evaluation metric. Nevertheless, this may not perform well since there exist just a few memories with positive reward signal throughout the experiment. Alternatively, we exploit the total discounted reward (return) or more precisely \(\sum_{i=t}^{T} \gamma^{i-t}r_i\). Having the return value, we can calculate the probability of forgetting the evaluated memory \((f)\):
\begin{equation}
f = e^{-\alpha(return)}	
\end{equation}
where \(\alpha\) is a hyper-parameter indicating how forgetful our replay memory is. Increasing the \(\alpha\) value leads to a less forgetful replay memory. In our experiments, we set \(\alpha=1\). To utilise this formula, we use a modified architecture of experience replay memory rather than its original version  \cite{lin1993reinforcement}. The original replay memory architecture enjoys a queue (FIFO) to store the memories and removes the memory at the head of the queue without any exception. However, RMR will examine the memory at the head location and calculates the probability of forgetting \(f\) it. Then, with probability of  \(f\) the memory is forgotten and with probability of \(1-f\) the memory is restored back to the tail of the queue (Fig.\ref{fig:rmr}). Below, Alg.1 \ref{rer_add} represents the Pseudo-code of the RMR's memory handling algorithm:\\

%
\begin{algorithm}[H]
\SetAlgoLined
\KwIn{new\_memory}
 oldest\_memory = queue.pop()\;
 \While{\(e^{-\alpha(oldest\_memory.return)} < random(0,1) \) }{
    queue.add(oldest\_memory)\;
    oldest\_memory = queue.pop()\;
 }
 queue.add(new\_memory)\;
 \caption{Adding a new memory to RMR's replay memory}
 \label{rer_add}
\end{algorithm}

\section{Experimental results}
We evaluated the performance of the proposed method in the parametrised HFO domain for the task of approaching the ball and scoring a goal. 
Three types of agents were considered as follows: 
\begin{itemize}
    \item The baseline agent which using epsilon-greedy with random action selection (DDPG).
    \item The baseline agent enhanced with curious exploration (DDPG + CE).
    \item The baseline agent enhanced with curious exploration and Return-based Memory Restoration (DDPG + CE + RMR)
\end{itemize}
We used binary reward formulation for all three types of the agents. For each agent, we conducted three independent train and test experiment. Each training and testing rounds take 50,000 and 1,000 episodes, respectively. Each training round took almost six hours on a Nvidia GeForce GTX 1070 GPU platform.\\

\begin{figure}[t!]
    \centering
    \vspace{-3mm}
    \includegraphics[width=8cm]{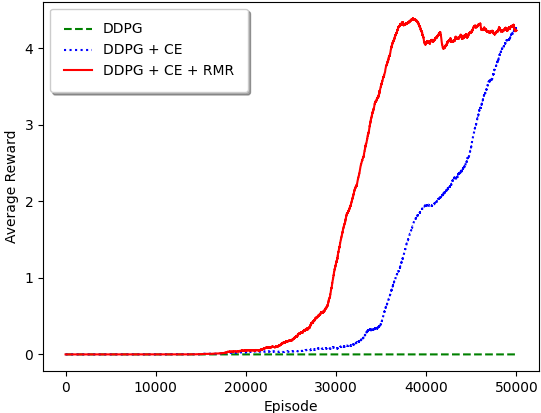}
    \vspace{-2mm}
    \caption{A comparison of average rewards among the three types of agents. The rewards are averaged over five rounds of experiments for each agent.}
    \label{fig:cmp_reward}
		\vspace{-3mm}
\end{figure}

The first agent, i.e. the standard DDPG failed to learn anything and never scored a single goal using binary reward function. Note that DDPG agent can perform the task with nearly-optimal performance using a hand-crafted reward function. In contrast, both DDPG + CE and DDPG + CE + RMR agents could reliably learn to score goals. It can be seen that DDPG + CE + RMR achieves a better performance in terms of the gained rewards per episode (Fig.\ref{fig:cmp_reward}). 

As represented in Fig.\ref{fig:cmp_reward}, DDPG + CE + RMR can converge faster to the optimal behaviour compared to DDPG + CE which is because of paying attention to more valuable and promising memories (maintained through RMR). In fact, the increased performance of DDPG + CE + RMR model is probably because of remembering earlier memories with positive return reward. Indeed, the agent without RMR fails to learn much in first episodes in which positive reward signal is seen due to the fact that it quickly forgets them. On the other hand, by using RMR, sparse positive signals become more dense in replay memory. However, when the positive reward signal becomes dense enough, the agent without RMR also learns how to score goals in our environment. In contrast, this may not be the case when the agent has to deal with a more complex environment. In other words, reward signal may stay sparse after exploration when the task is hard to be learned. 

As mentioned before, we ran 5 rounds of experiment (training and testing) for each type of agent and calculated the success rate per round. Table \ref{tbl:results} shows the success rate achieved by each agent in each round as well as the average result acquired by the agent. According to the results, DDPG completely fails to do the task; DDPG + CE can achieve \%91.9 success rate and DDPG + CE + RMR outperform both other methods by almost a margin of \%6. 

\begin{table}[t]
\centering
\vspace{2mm}
\begin{tabular}{c|c|c|c}
       & \textbf{ DDPG }  & \textbf{ DDPG + CE } & \textbf {DDPG + CE + RMR } \\ \hline
Round 1 & 0.000 & 0.978     & 0.989           \\
Round 2 & 0.000 & 0.977     & 0.988           \\
Round 3 & 0.000 & 0.917     & 0.986           \\
Round 4 & 0.000 & 0.867     & 0.964           \\
Round 5 & 0.000 & 0.858     & 0.963           \\ \hline
Avg    & 0.000 & 0.919     & \textbf{0.978}
\end{tabular}
\vspace{2mm}
    \caption{The success rate achieved by the agents examined in our experiments. As expected, DDPG + CE + RMR agent shows the best performance among the three.}
    \label{tbl:results}
    \vspace{-8mm}
\end{table}

\begin{figure}[t!]
    \centering
    \vspace{-3mm}
    \includegraphics[width=11.5cm]{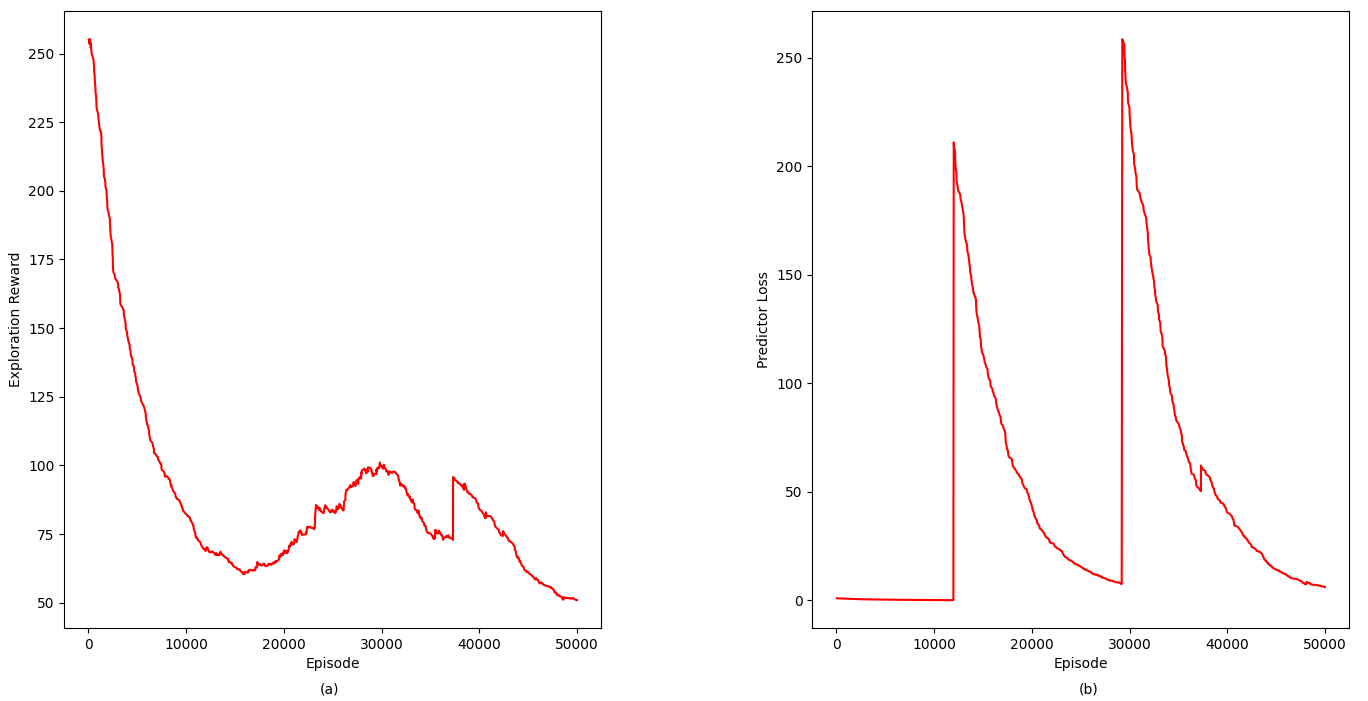}
    \vspace{-3mm}
    \caption{a) Exploration reward graph gained by the exploration agent; b) Prediction loss graph that shows two spikes revealing the time that the agent first intercepts the ball and when the agent scores a goal for the first time}
    \label{fig:loss_rew}
		\vspace{-1mm}
\end{figure}

For a further analysis on Curious Exploration, the predictor's loss and  explorer's reward gain are plotted (Fig. \ref{fig:loss_rew}). As can be seen, higher reward gain of the explorer (Fig.\ref{fig:loss_rew}.a) corresponds to the higher prediction error (Fig.\ref{fig:loss_rew}.b). The first spike in predictor's loss (after learning basic dynamics) happens when the agent starts interacting with the ball. After noticing a strange behaviour from the ball (e.g. when the ball is kicked), the agent becomes extremely curious about this object. By interacting and examining this object while trying to learn its dynamics, it will eventually score goals. Now, because it did not know about positive reward signal gained by scoring a goal, it becomes curious about scoring. Through this process, the amount of exploration is annealed towards exploitation. 
\vspace{-1mm}
\section{Conclusion and Future Work}
\vspace{-1mm}
This paper presented a framework capable of learning how to score on an empty goal in HFO domain with only a binary success/failure reward using a combination of DDPG, Curious Exploration, and Return-based Memory Restoration methods. The research showed how the training process can be improved by exploiting Return-based Memory Restoration which seeks to remember more valuable memories. The experimental results confirmed our proposed method can achieve nearly-optimal behaviour while the baseline method entirely failed to learn any tasks. 

The prediction-based exploration method, used in this article, is prone to noisy-TV problem. In our environment, however, we did not face such an issue; but, this may change with more complex situations where it may include stochastic agents. To deal with this challenge, we suggest applying ICM \cite{pathak2017curiosity} to stabilise the exploration process.

Furthermore, the behaviour of RMR must be further examined in more complicated environments. An interesting possibility is to use RMR (which modifies the memory maintenance procedure in experience replay memory) along with a state-of-the-art sampling method such as Prioritized Experience Replay (PER). The learning performance of the method could be also investigated for more complex tasks using only a binary success/failure reward function.

\bibliographystyle{unsrt}  
\bibliography{template.bib}

\end{document}